\relax
\documentclass[letterpaper]{article} % DO NOT CHANGE THIS
\usepackage{aaai22}  % DO NOT CHANGE THIS
\usepackage{times}  % DO NOT CHANGE THIS
\usepackage{helvet}  % DO NOT CHANGE THIS
\usepackage{courier}  % DO NOT CHANGE THIS
\usepackage[hyphens]{url}  % DO NOT CHANGE THIS
\usepackage{graphicx} % DO NOT CHANGE THIS
\usepackage{natbib}  % DO NOT CHANGE THIS AND DO NOT ADD ANY OPTIONS TO IT
\usepackage{caption} % DO NOT CHANGE THIS AND DO NOT ADD ANY OPTIONS TO IT
\DeclareCaptionStyle{ruled}{labelfont=normalfont,labelsep=colon,strut=off} % DO NOT CHANGE THIS
\usepackage{graphicx}
\usepackage{amsmath}
\usepackage{amsthm}
\usepackage{booktabs} 
\usepackage{amsfonts,amssymb}
\usepackage{amsmath}
\usepackage{threeparttable}
\usepackage{subfigure}
\usepackage{booktabs}
\usepackage{color}
\usepackage{algorithm}
\usepackage{algorithmic}
\usepackage{times}
\usepackage{soul}
\usepackage{url}

\usepackage{newfloat}
\usepackage{listings}
\floatstyle{ruled}
\newfloat{listing}{tb}{lst}{}
\floatname{listing}{Listing}
\title{Deep Unsupervised Active Learning on Learnable Graphs}
\author{
    Handong Ma\textsuperscript{\rm 1},
    Changsheng Li\textsuperscript{\rm 2}, 
    Xinchu Shi\textsuperscript{\rm 3}, 
    Ye Yuan\textsuperscript{\rm 2}, 
    Guoren Wang\textsuperscript{\rm 2}
}
\affiliations {
    \textsuperscript{\rm 1} SCSE, University of Electronic Science and Technology of China, Chengdu, China\\
    \textsuperscript{\rm 2} School of Computer Science and Technology, Beijing Institute of Technology, Beijing, China\\
    \textsuperscript{\rm 3} Meituan \\
    handongma@std.uestc.edu.cn, \{lcs, yuan-ye\}@bit.edu.cn, shixinchu@meituan.com, wanggrbit@126.com
}

\usepackage{bibentry}
\begin{document}

\maketitle

\begin{abstract}
Recently deep learning has been successfully applied to unsupervised active learning. 
However, current method attempts to learn a nonlinear transformation via an auto-encoder while ignoring the sample relation, leaving huge room to design more effective representation learning mechanisms for unsupervised active learning.
In this paper, we propose a novel deep unsupervised Active Learning model via Learnable Graphs, named ALLG.
ALLG benefits from learning optimal graph structures to acquire better sample representation and select representative samples.
To make the learnt graph structure more stable and effective, we take into account $k$-nearest neighbor graph as a priori, and learn a \textit{relation propagation} graph structure.
We also incorporate shortcut connections among different layers, which can alleviate the well-known over-smoothing problem to some extent.
To the best of our knowledge, this is the first attempt to leverage graph structure learning for unsupervised active learning.
Extensive experiments performed on six datasets demonstrate the efficacy of our method.
\end{abstract}

\section{Introduction}
Active learning is an active research topic in machine learning and computer vision communities.
Its goal is to choose informative or representative samples to be labeled, so as to reduce the costs of annotating but guarantee the performance of the model trained on these labeled samples.
Due to its huge potentiality, active learning has been widely applied to various tasks, such as image classification \cite{wang2016cost}, recommendation systems \cite{elahi2016survey}, object detection \cite{Aghdam_2019_ICCV}, semantic segmentation \cite{siddiqui2020viewal} and so on.

%According to whether involving labeled data during sample selection, active learning can be roughly categorized into two groups: supervised and unsupervised ones.
%Supervised active learning methods usually pre-train one or several models with limited labeled data to iteratively select informative samples according to some criteria, such as uncertainty, diversity, etc \cite{macskassy2009using,Jain2009Active,Huang2010Active,Elhamifar2013A,Zheng2015Querying,Zhang2017Bidirectional,haussmann2019deep}. 
%Many supervised active learning methods \cite{macskassy2009using,Jain2009Active,Huang2010Active,Elhamifar2013A,Zheng2015Querying,Zhang2017Bidirectional,haussmann2019deep} have been proposed in the past decades.
%Since labeled data participates in sample selection, supervised active learning methods can achieve good results but with a relatively high complexity of model training.

Unsupervised active learning targets at selecting representative samples through taking advantage of structure information of data. 
Currently, most unsupervised approaches \cite{yu2006active,zhang2011active,zhu201510,shi2016diversifying,li2018joint} intend to minimize the data reconstruction loss with different structure regularization terms for selecting representative samples.
These methods assume that each data point can be represented by a linear combination of a selected sample subset, thus failing in modeling data with nonlinear structures.
To remedy this issue, a kernel based method \cite{cai2011manifold} incorporates the manifold structure of data into the reproducing kernel Hilbert space (RKHS).
More recently, deep learning has been applied to solve the unsupervised active learning problem \cite{liijcai20}, named DUAL. DUAL attempts to nonlinearly map data into a latent space, and then performs sample selection in the learnt space.

The key to success for DUAL stems from learning a nonlinear transformation to obtain new feature representations by leveraging deep learning. 
However, it does not explicitly take advantage of the relation among samples during representation learning. 
Recently, graph neural networks (GNN) have attracted much attention \cite{kipf2016semi,velivckovic2017graph,xu2018powerful}, where the sample relation has been proved to be helpful for learning good sample representations. Thus, it ought to be beneficial to sample selection, if we can leverage graph structure information of data, and aggregate neighbor information of samples to learn better representations. But another question arises: how to construct an optimal graph structure for non-graph data still remains an open problem.

%Although DUAL can model data with nonlinear structures via a deep auto-encoder and learn a distinct embedding for each data point depending on the correlations between features, it ignores the relation among samples.
%In many scenarios, the feature of a single data point may be hard to capture the complete properties of itself.
%So sample selection conducted by those general approaches may get sub-optimal results.
%Recently, graph neural networks (GNN) have attracted much attention \cite{wu2020comprehensive}, where one node aggregates features from its local neighbor. 
%Typical work including \cite{kipf2016semi,velivckovic2017graph,xu2018powerful} show great power of GNN as samples can obtain complementary information from their "neighbors" to gain more precise representation.
%Thus, it will be beneficial to sample selection, if we can leverage graph structure to learn a better sample representation.
%But another question has been arisen here that how to define an optimal graph structure for non-graph data, as most GNN models require pre-define intrinsic graph structure (e.g., social network) of data.

\begin{figure*}[ht]
\centering
\includegraphics[width=1\linewidth]{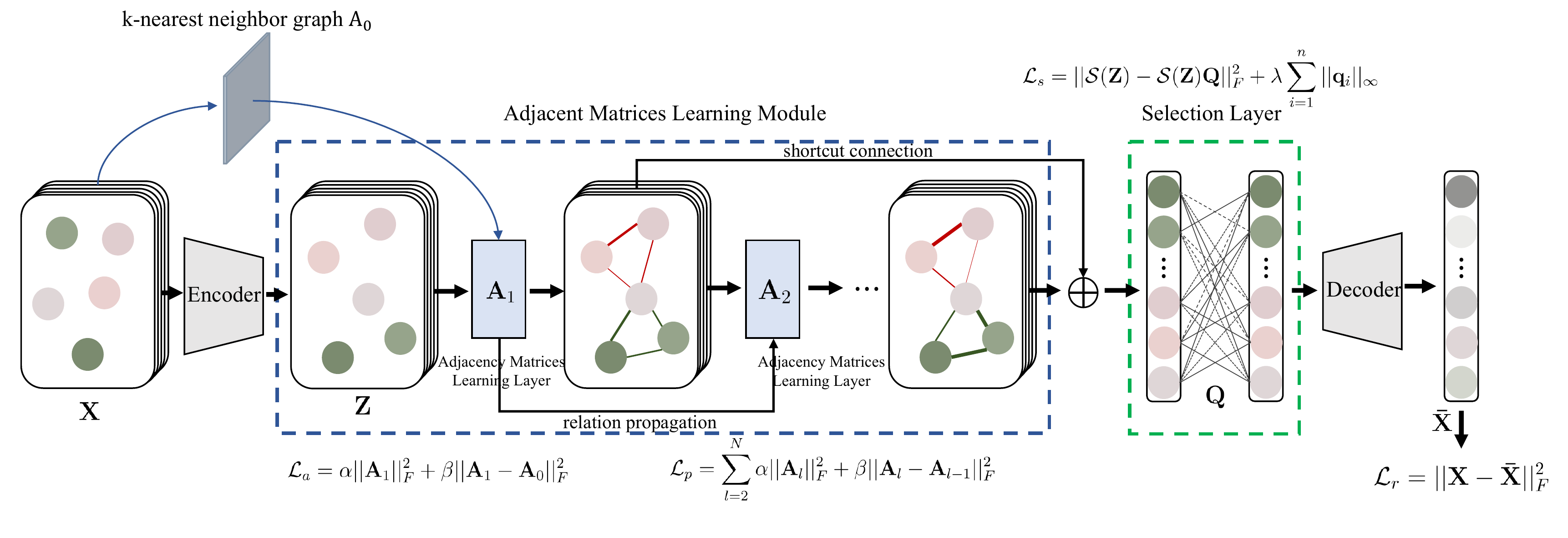}
\caption{The overall framework of ALLG. It mainly consists of four modules: An encoder and a decoder are used to learn a nonlinear mapping. An adjacent matrix learning module aims to learn a relation propagation regularized graph structure for precise sample representation. 
A shortcut connection is introduced to alleviate the over-smoothing problem. A self-selection layer is used to select representative samples.}
\label{framework}
\end{figure*}

Based on the above considerations, we propose a novel deep unsupervised Active Learning model based on Learnable Graphs, called ALLG. 
%The overall framework of ALLG is presented in Figure\ref{framework}. 
Specifically, ALLG first utilizes an auto-encoder framework to map samples into a latent space.
Without pre-defining graph structure of samples, ALLG devises a novel adjacent matrices learning module to automatically learn an optimal graph structure among samples for jointly refining sample representations and sample selection. 
In this module, we incorporate the $k$-nearest neighbor graph as a priori to learn a stable graph structure. Considering that the relation among samples may happen to evolve as the sample representations change in different network layers, we attempt to learn a series of relation propagated adjacent matrices, in the hope of capturing more precise graph structures.
Moreover, we add shortcut connections among different adjacent matrices learning layers, which can alleviate the well-known over-smoothing problem to some extent.
Finally, a self-selection layer is employed to select representative samples based on the learnt sample representation.

%We briefly summarize the contributions of this paper:
The contributions of this paper are summarized as:
\begin{itemize}
\item ALLG builds a connection between unsupervised active learning and graph structure learning. To the best of our knowledge, this is the first attempt to leverage graph structure learning for unsupervised active learning.
\item ALLG attempts to learn more precise sample representation by leveraging the graph, and devise a novel mechanism to dynamically learn a series of graph adjacent matrices.
\item Inspired by the idea of residual learning, ALLG adds shortcut connections among different adjacent matrices learning layers to alleviate the over-smoothing problem.
\end{itemize}

Extensive experiments are performed on six publicly available datasets, and experimental results demonstrate the effectiveness of ALLG, compared with the state-of-the-arts.

\section{Related Work}
In this section, we will briefly review some work on unsupervised active leaning and graph structure learning.
\subsection{Unsupervised Active Learning}
Unsupervised active learning has attracted much attention in recent years. 
At the earlier stage, \cite{yu2006active} propose to utilize transductive experimental design (TED) to select a sample subset, and obtain a greedy solution.
ALNR \cite{hu2013active} performs sample selection by considering the neighborhood relation of samples.
RRSS \cite{Nie2013Early} proposes a convex formulation by introducing a structured sparsity-inducing norm, and a robust sparse representation loss. 
To select complementary samples, \cite{shi2016diversifying} proposes a diverse loss function that is an extension of TED.
LSR \cite{li2017active} was proposed via local structure reconstruction to select representative data points.
ALFS \cite{li2018joint} builds a connection between unsupervised active learning and feature selection, and proposes a convex formulation to select samples and features simultaneously.

Most recently, owing to the powerful representation capability and great success of deep learning, deep model has been explored to solve the unsupervised active learning problem \cite{liijcai20}. In \cite{liijcai20}, the authors utilize deep auto-encoders to embed data to a latent space, and then select the most representative samples to best reconstruct both the whole dataset and clustering centroids. 
Although deep learning based methods have achieved impressive results, they ignore the relation among data points during sample representation learning and thus have the inferior results.  

\subsection{Graph Structure Learning}
For the sake of applying GNNs to non-graph structured data, many graph structure learning methods have been proposed in recent years.
\cite{dong2016learning,egilmez2017graph} explore to learn the graphs from data without associating it with the downstream tasks. 
More recently, \cite{li2018adaptive,choi2019graph,liu2019contextualized,chen2019deep} aim to dynamic construct graphs towards the downstream tasks. 
However, these methods are task-specific ones which depend on the supervised information.

Different from these models, we propose to optimize the learning of graphs and active learning simultaneously in an unsupervised manner.

\section{Method}
The overall architecture of our method is illustrated in Figure \ref{framework}.
Our network mainly consists of the following components: an encoder and decoder module is used to learn a non-linear transformation.
An adjacent matrix learning module aims to learn multiple optimal adjacent matrices and leverage them for learning compact sample representation, which is the core module of our method.
A self-selection module attempts to select representative samples.
Before introducing these modules in detail, we first give some notations.

Let $\mathbf{X}=[\mathbf{x}_1, \mathbf{x}_2, \cdots, \mathbf{x}_n] \in \mathbb{R}^{d \times n}$ denote a data matrix, where $\mathbf{x}_i (1\textless i\textless n)$ is the data point. 
$d$ and $n$ are the dimension and number of data points respectively. 
Our goal is to learn a nonlinear transformation by considering the sample relation to learn better sample representations. 
% After that, we intend to select $m$ representative samples to best approximate the learnt new representation.
To this end, we first learn a latent space with a deep auto-encoder. 
We then attempt to learn the graph structure of the data in the latent space, and leverage it for learning a good representation. 
Based on the learnt representation, we can select the most representative samples via a self-selection layer. 
We attempt to optimize them in a joint framework.

\subsection{Encoder and Decoder}
In order to learn a nonlinear transformation $\Theta$, we utilize an auto-encoder architecture to map the data into a latent space, because of its effectiveness in unsupervised learning.

In our framework, the encoder consists of $L$ fully connected layers. 
The output of the $l$-th layer in the encoder is defined as:
\begin {equation}
\mathbf{z}_i^{(l)}=\sigma(\mathbf{W}^{(l)}\mathbf{z}_i^{(l-1)}+\mathbf{b}^{(l)}), \qquad i=1, ... , n
\end {equation}
where $\mathbf{z}_i^{(0)}=\mathbf x_i$ denotes the $i$-th original training data in $\mathbf X$, which is used as the input of the encoder block. 
$\mathbf{W}^{(l)}$ and $\mathbf{b}^{(l)}$ are the weights and bias associated with the $l$-th hidden layer respectively. 
$\sigma(\cdot)$ is a nonlinear activation function. Then, we can define the latent representation as:
\begin {equation}
\mathbf Z^L=\Theta(\mathbf X)=[\Theta(\mathbf x_1), \Theta(\mathbf x_2), \cdots, \Theta(\mathbf x_n)] \in \mathbb{R}^{d' \times n}
\end {equation}
where $d'$ denotes the dimension of the latent representation.

As for the decoder, it learns another nonlinear mapping to reconstruct the original data, which guides the training of the encoder. 
The decoder has a symmetric structure with the encoder, which consists of $L$ fully connected layers as well. 

Then, the reconstruction loss of auto-encoder is defined as:
\begin{equation}\label{reconstruction}
\mathcal{L}_r = \sum_{i=1}^n||\mathbf{x}_i - \mathbf{\bar{x}}_i||^2_2 = ||\mathbf{X} - \mathbf{\bar{X}}||_F^2,
\end{equation}
where $\mathbf{\bar{X}}$ denotes the reconstruction of $\mathbf{X}$.
Actually, the input $\mathbf{X}$ plays a role of self-supervisor to guide the learning of auto-encoder.

\subsection {Graph Structure Learning}
After obtaining the latent representation $\mathbf{Z}^L$ of the input $\mathbf{X}$, we intend to learn the graph structure of the input $\mathbf{X}$ for generating more precise sample representation. 

% As aforementioned, most GNN algorithms require to pre-define intrinsic graph structure of data. 
% For non-graph data, human estimated graph structures such as $k$-nearest neighbor graph can not guarantee the optimal of it.

\vspace{0.1in}
\noindent \textbf{Adjacent Matrix Learning}: 
As aforementioned, taking advantage of the relation among data points can have a positive effect on sample representation learning, which has been verified in graph neural networks (GNNs) \cite{wu2020comprehensive}. 
However, many GNN algorithms are developed to deal with graph data, and assume that the adjacent relationship is given, which can not be directly applied to non-graph data.
To deal with it, there are some algorithms which attempt to construct a human estimated graph structure, e.g., $k$-nearest neighbor graph.
However, such methods can not guarantee the graph structure is optimal. 
Based on these considerations, we aim to learn a data-driven optimal graph structure.
Considering that some human estimated graph structure can still reveal some priori information, we can integrate such information into our framework to regularize the graph structure learning.
In all, we propose the following regularization terms to learn the graph structure: 
\begin {equation} \label{matrixlearning}
\mathcal{L}_a= \alpha||\mathbf{A}_1||_F^2+\beta||\mathbf{A}_1-\mathbf{A}_0||_F^2 ,
\end {equation}
where 
$\mathbf A_0$ denotes the $k$-nearest neighbor graph, and $\mathbf{A}_1$ denotes the learnt graph, i.e., the adjacent matrix, $\alpha$ and $\beta$ are positive trade-off parameters.

In Eq.(\ref{matrixlearning}), the first regularization term aims to reduce the complexity of the learnt adjacent matrix.
The second term imposes a constraint on $\mathbf{A}_1$, making it not deviate from the $k$-nearest neighbor graph $\mathbf{A}_0$, such that the prior information can be integrated. 
One can control how close the learnt adjacent matrix to the priori by modifying the parameter $\beta$.
Actually, the graph adjacent matrix $\mathbf A_1$ can be regarded as the parameters of a fully connected layer without bias. Therefore, it can be updated through a standard back-propagation procedure during training.

After obtaining the adjacent matrix $\mathbf A_1$, we can obtain the new sample representations $\mathbf S_1$ based on $\mathbf Z^L$ as:
\begin{align}
    \mathbf {S}_1 = \sigma({\mathbf {Z}^L \mathbf {A}_1}) ,
\end{align}
where $\sigma (\cdot)$ is a nonlinear activation function.
Based on the above equation, we can see that the new sample representation $\mathbf S_1$ can be obtained based upon a linear combination of $\mathbf {Z}^L$ with the weight matrix $\mathbf {A}_1$, thereby the relation among samples can be incorporated into the process of representation learning. 

\vspace{0.1in}
\noindent \textbf{Relation Propagation}:
It is worth noting that only learning one adjacent matrix may be sub-optimal for learning the graph structure of data, since the representations of samples are from different layers during the process of training.
In order to learn more stable and effective optimal graph structure, we intend to learn multi-level adjacent matrices with a relation regularized term.
In order to enable the learning of multiple adjacent matrices, we propose another regularization terms as:
\begin {equation}\label{smoothing}
\mathcal{L}_{p}=\sum_{l=2}^{N} (\alpha'||\mathbf{A}_l||_F^2+\beta'||\mathbf{A}_l-\mathbf{A}_{l-1}||_F^2),
\end {equation}
where $\mathbf{A}_l$ denotes the learnt adjacent matrix of the $l$-th matrix learning layer, $\alpha'$ and $\beta'$ are positive trade-off parameters.

In Eq.(\ref{smoothing}), the first term aims to lower the complexities of the learnt adjacent matrices, while the second term utilizes the former learnt adjacent matrix to regularize the latter adjacent matrix. 
By this means, the relation among samples can be smoothly propagated. 

After learning multiple adjacent matrices, we can obtain another sample representation for sample selection as:
\begin{align}
    \mathbf {S}_{i+1} = \sigma({\mathbf {S}_i \mathbf {A}_i}), i=1,2,\cdots, N-1
\end{align}
where $\mathbf{S}_{i}$ denotes the latent sample representations in the $i$-th graph Laplacian layer. 

\begin{figure}[h!]
\centering
\includegraphics[width=0.9\linewidth]{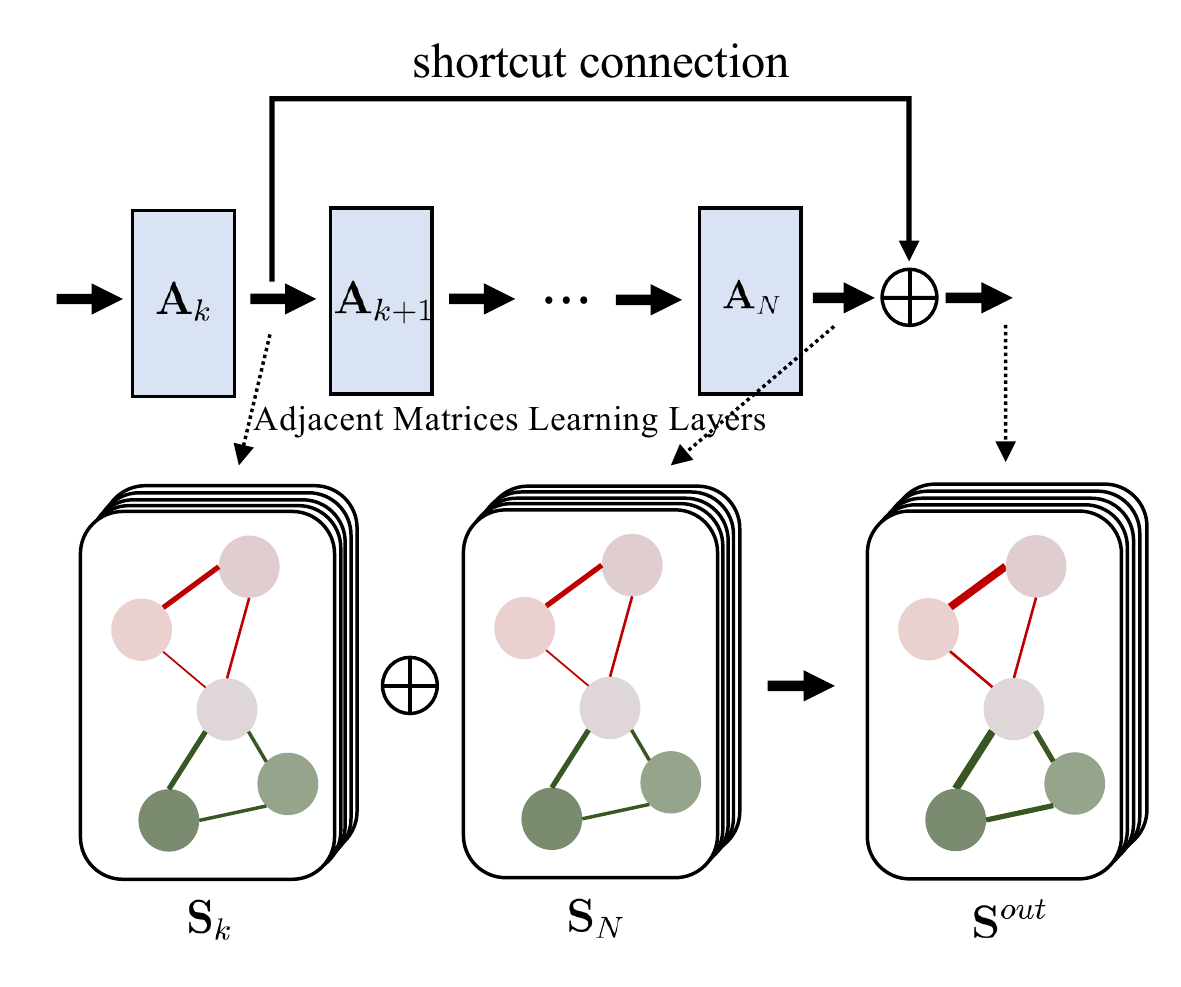}
\caption{The short connection in ALLG}
\label{shortcutfig}
\end{figure}

\vspace{0.1in}
\noindent \textbf{Shortcut Connection}:
In GNNs, there is a well known problem, i.e., over-smoothing. The over-smoothing problem means that repeated graph Laplacian eventually make node embeddings indistinguishable. Empirically, a shortcut connection would bring more discriminative features from the former layers to alleviate this problem. 
The shortcut connection is illustrated in Figure \ref{shortcutfig}.
Mathematically, it can be described as:
\begin {equation}\label{shortcut}
\mathbf{S}^{out}= r \mathbf{S}_k + (1-r)\mathbf{S}_N,
\end {equation}
where $r$ is a trade-off parameter to control the contribution of the $k$-th graph structure learning layer to the final sample representation.

\subsection{Self-Selection Layer}
After obtaining the final latent sample representations, we perform sample selection by introducing a self-selection layer, as shown in Figure \ref{framework}.

Self-selection layer takes the output of the adjacent matrix learning layers as the input.
In order to select samples to best reconstruct all ones, we use the loss function presented as:
\begin{equation}\label{nphard}
\begin{split}
\mathcal{L}_s=\sum_{l=1}^{n} ||\mathbf{S}^{out}-\mathbf{S}^{out}\mathbf{Q}||_2^2+\lambda ||\mathbf{Q}||_{\infty,1}
%\\s.t.\ \ \mathcal{S}(\mathbf Y)=[\mathcal{S}(\mathbf y_1), \mathcal{S}(\mathbf y_2) ,..., \mathcal{S}(\mathbf y_m)] \in \mathbb{R}^{d' \times m}
\end{split}
\end{equation}
where  $\mathbf{Q} \in \mathbb{R}^{n\times n}$ is the reconstruction coefficients for the samples.
$||\mathbf{Q}||_{\infty,1}= \sum_{i=1}^{n}||\mathbf{q}_{i}||_\infty$, where $\mathbf{q}_{i}$ is the $i$-th row vector of $\mathbf{Q}$ and $||\cdot||_\infty$ denotes the sup-norm of a vector, defined as $||\mathbf{a}||_{\infty}=\max \limits_{1\leq i \leq n} |a_i|$.

The first term in Eq. (\ref{nphard}) aims to pick out $m$ samples to reconstruct the whole dataset in the latent space, while the second term is a regularization term to enforce the coefficient matrix $\mathbf{Q}$ row-sparse.
To minimize Eq. (\ref{nphard}), similarly, $\mathbf Q$ can be regarded as the parameters of a fully connected layer without bias and nonlinear activations, and solved jointly through a standard back-propagation procedure.

After the inputs passing by the self-selection layers, we then feed them into the decoder as its inputs as:
\begin{align}
    \mathbf{\bar X}_{0}=\mathbf S_{out} \mathbf{Q} \in \mathbb{R}^{d' \times n}
\end{align}

\subsection{Overall Model and Training}

Based on Eq. (\ref{reconstruction}), (\ref{matrixlearning}), (\ref{smoothing}), and (\ref{nphard}),  the final loss function is
\begin{equation}\label{totalloss}
\mathcal L_{total}=\mathcal L_r+\mathcal L_{a}+\mathcal L_{p}+ \mathcal L_{s}
 \end{equation}

To jointly optimize Eq.(\ref{totalloss}), we use a two-stage training strategy following \cite{liijcai20}:
Firstly, we only pre-train the encoder-decoder without considering the matrices learning module and self-selection layer, minimizing the loss in Eq. (\ref{reconstruction}). By this means, we obtain good initial parameters for fine-tuning the whole model. Specifically, three fully connection layers are used in the encoder, and the decoder has a symmetric structure with it.
After that, we update all the parameters by minimizing Eq.(\ref{totalloss}) through a standard back-propagation procedure.
The ReLU is used in our method as the activations. And we optimize Eq. (\ref{totalloss}) with the Adam optimizer where we set the learning rate to $1.0\times10^{-3}$. In the final model, ALLG utilizes 2 adjacent matrices learning layers. As for the shortcut connection, we set the $k=1$ and $r=0.3$ in Eq.(\ref{shortcut}) during training. The hyperparameters $\alpha$ and $\alpha'$ are set to be the same so as the $\beta$ and $\beta'$ during training.
The optimization procedure for ALLG can also be seen in Algorithm.\ref{allg:algorithm}

\subsection{Post Processing} Once the model converges, we can get the parameter $\mathbf Q$ based on the self-selection layer. As $\mathbf Q$ is row sparse after training, the row values of $\mathbf Q$ can be regarded as the contributions of each data point to reconstruct other data points. Thus, we calculate the $\ell_2$-norm of the rows and sort them in descending order to get a rank of samples. Then we can select the top-$m$ data points as the most representative samples to be labeled.

\begin{algorithm}[t]
\caption{Optimization Procedure for ALLG}
\label{allg:algorithm}
\textbf{Input}: The matrix $\mathbf{X} \in \mathbb{R}^{d \times n}$\\
\textbf{Parameter}: trade-off parameters $\alpha,\ \beta,\ \lambda$\\
\textbf{Output}: $\mathbf{Q}$
\begin{algorithmic}[1] %[1] enables line numbers
\STATE Calculate the $k$-nearest neighbor graph $\mathbf{A}_0$ with $\mathbf{X}$.
\STATE Pre-train an auto-encoder network which has the same architecture with the encoder and decoder block of the proposed framework via a standard back-propagate algorithm on $\mathbf{X}$.
\STATE Initialize the encoder and decoder block of the proposed framework using the pre-trained network parameters.
\WHILE{not converged}
\STATE Update the whole network by minimizing the loss function in Eq.(\ref{totalloss}).
\ENDWHILE
\end{algorithmic}
\end{algorithm}

\section{Experiments}
In this section, we will introduce the experimental results to demonstrate the effectiveness of the proposed method.
We also conduct some ablation study and analysis on the proposed ALLG.

\begin{figure*}[t!]
\centering
%\subfigtopskip=0.5pt %设置子图与上面正文或别的内容的距离
%\subfigbottomskip=0.5pt %设置第二行子图与第一行子图的距离，即下面的头与上面的脚的距离
%\subfigcapskip=-3pt %设置子图与子标题之间的距离
\subfigure[Splice-junction]{\includegraphics[width=2.1in]{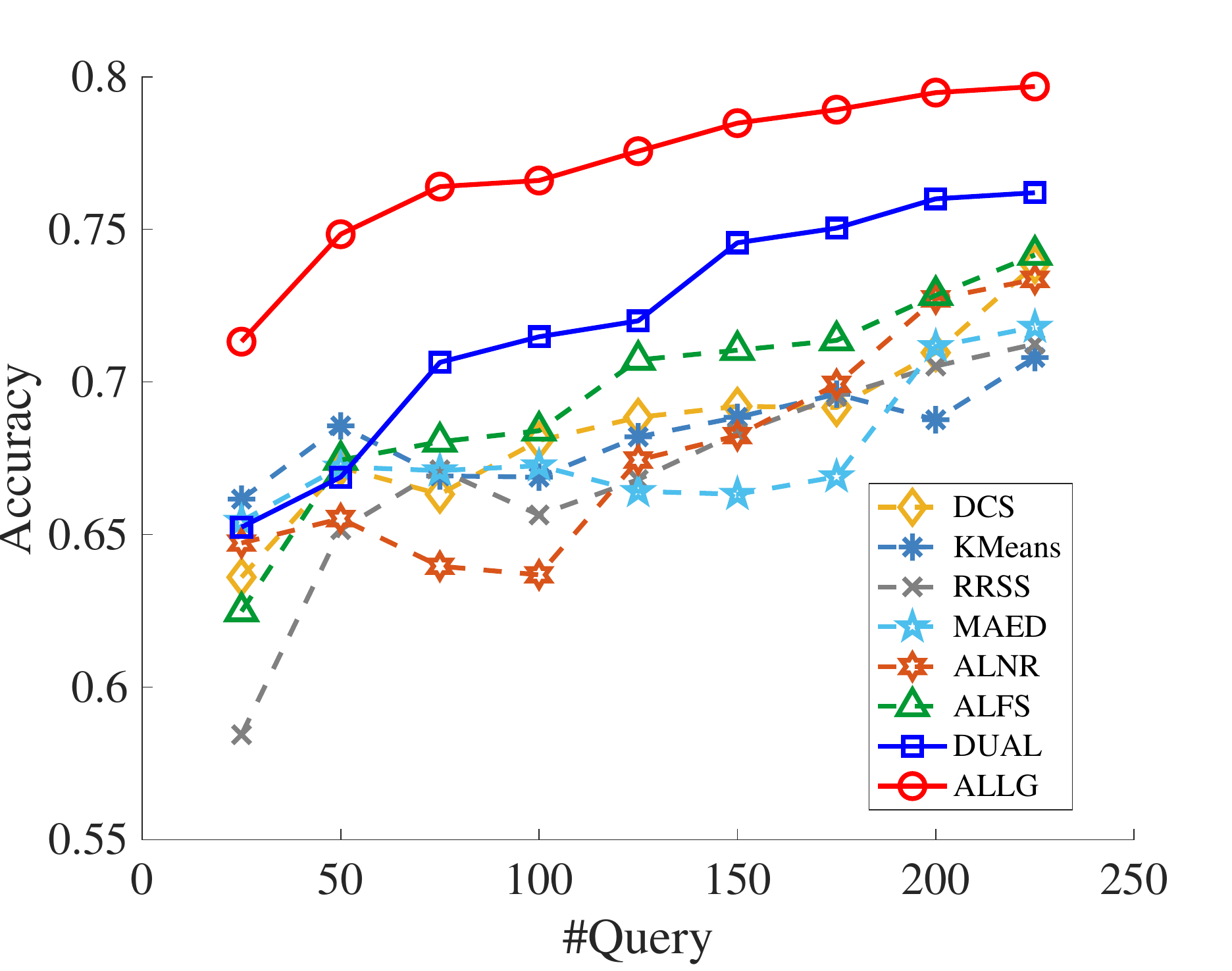}}
\subfigure[Plant Species Leaves]{\includegraphics[width=2.1in]{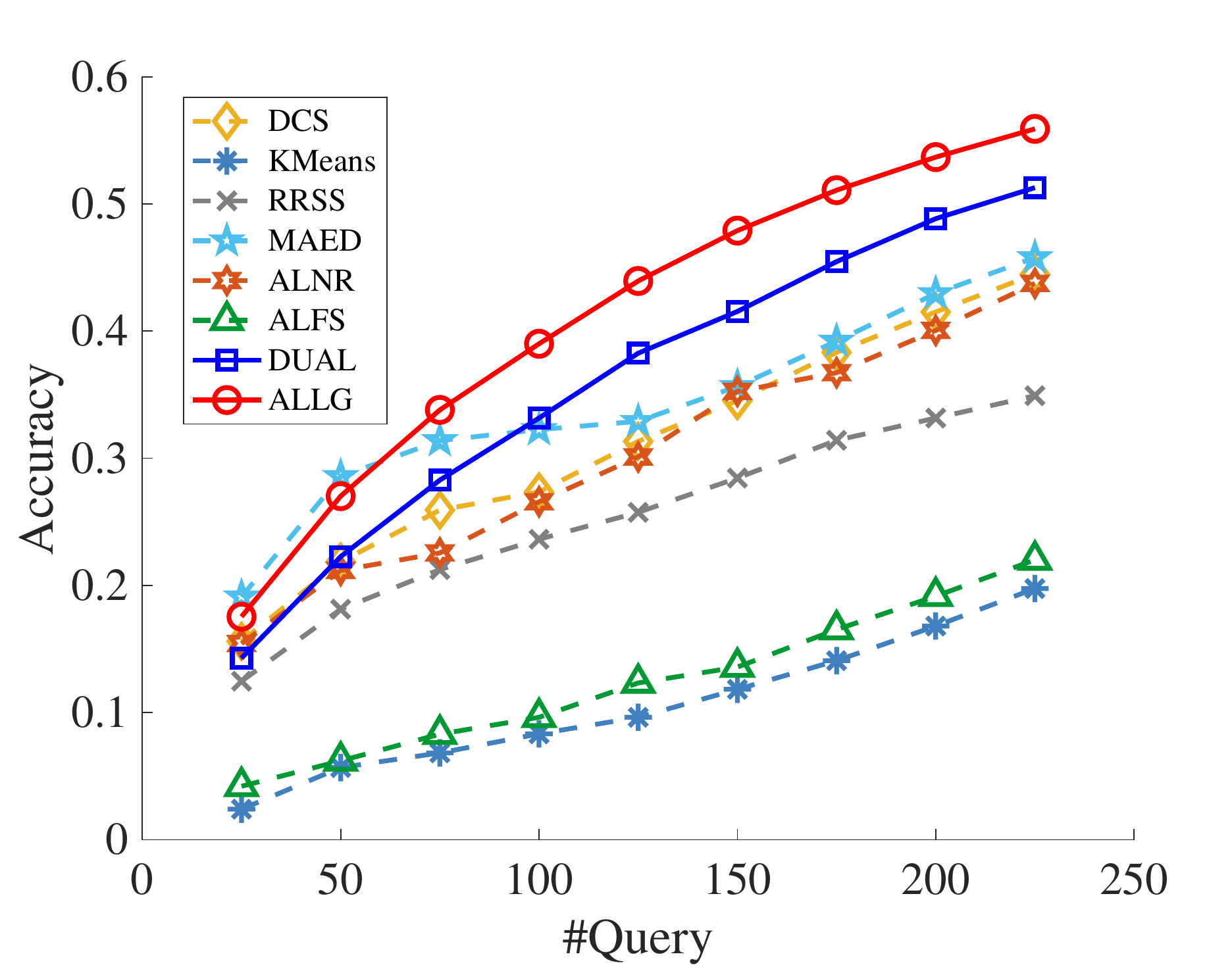}}
\subfigure[Waveform]{\includegraphics[width=2.1in]{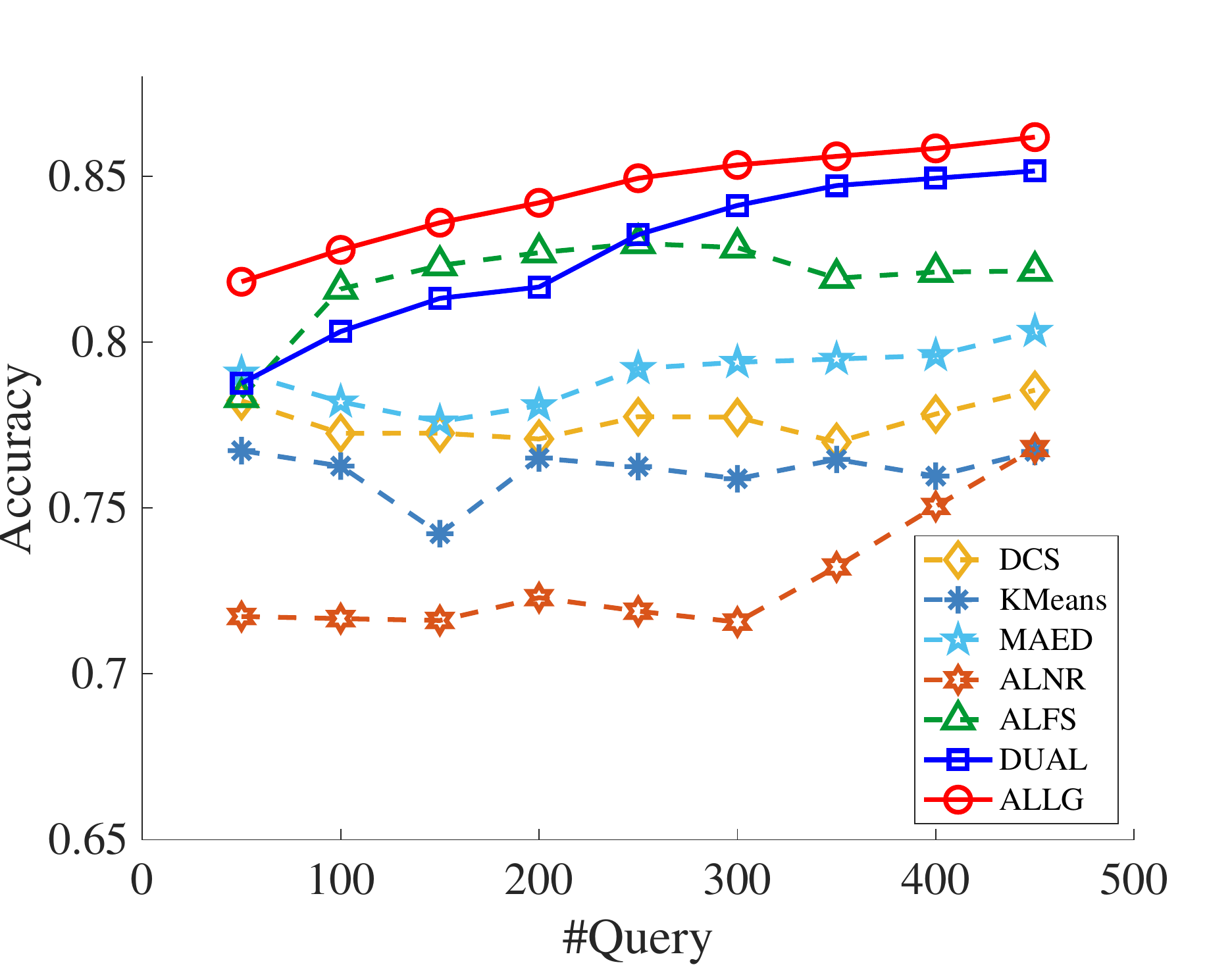}}
\subfigure[ESR]{\includegraphics[width=2.1in]{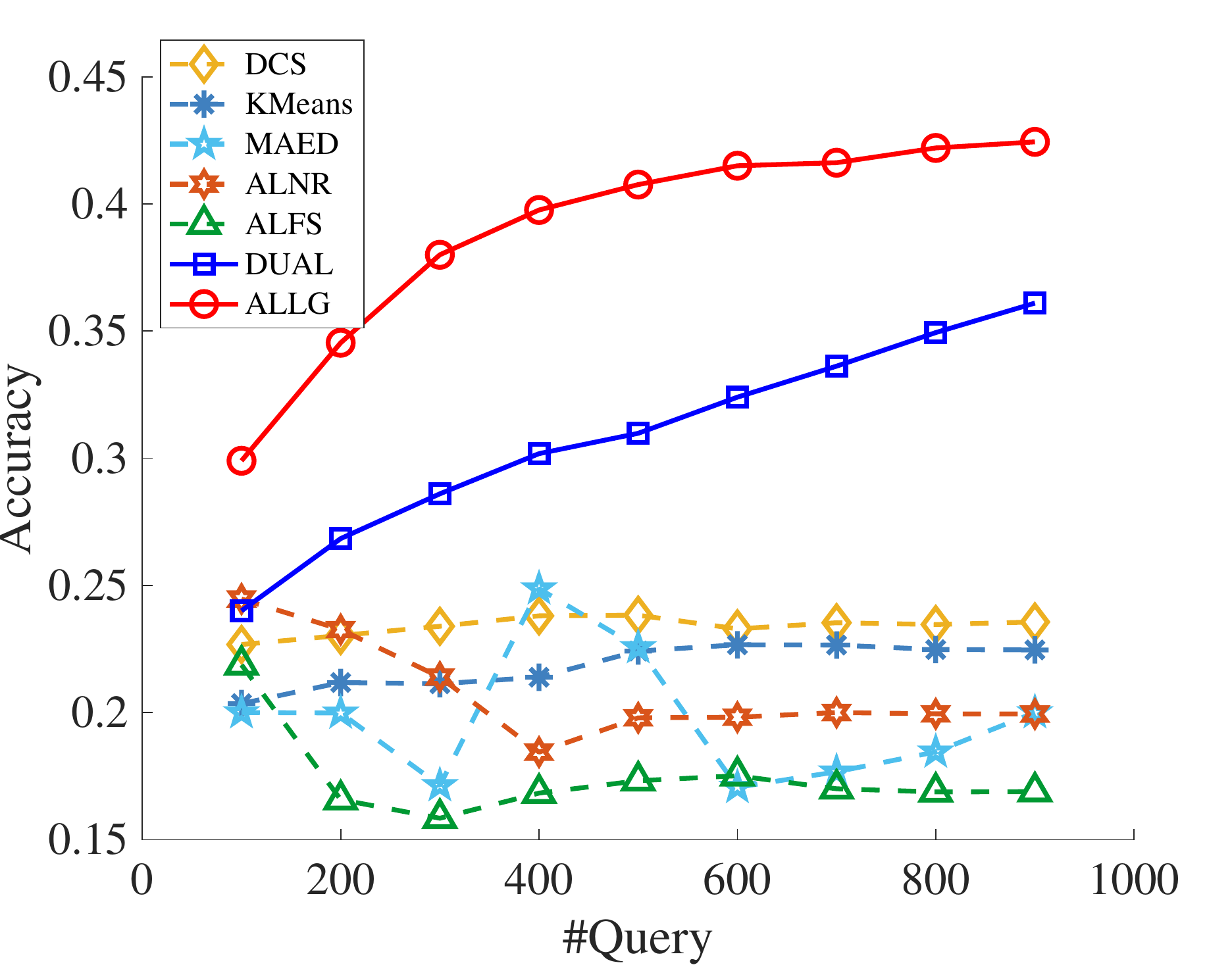}}
\subfigure[GSAD]{\includegraphics[width=2.1in]{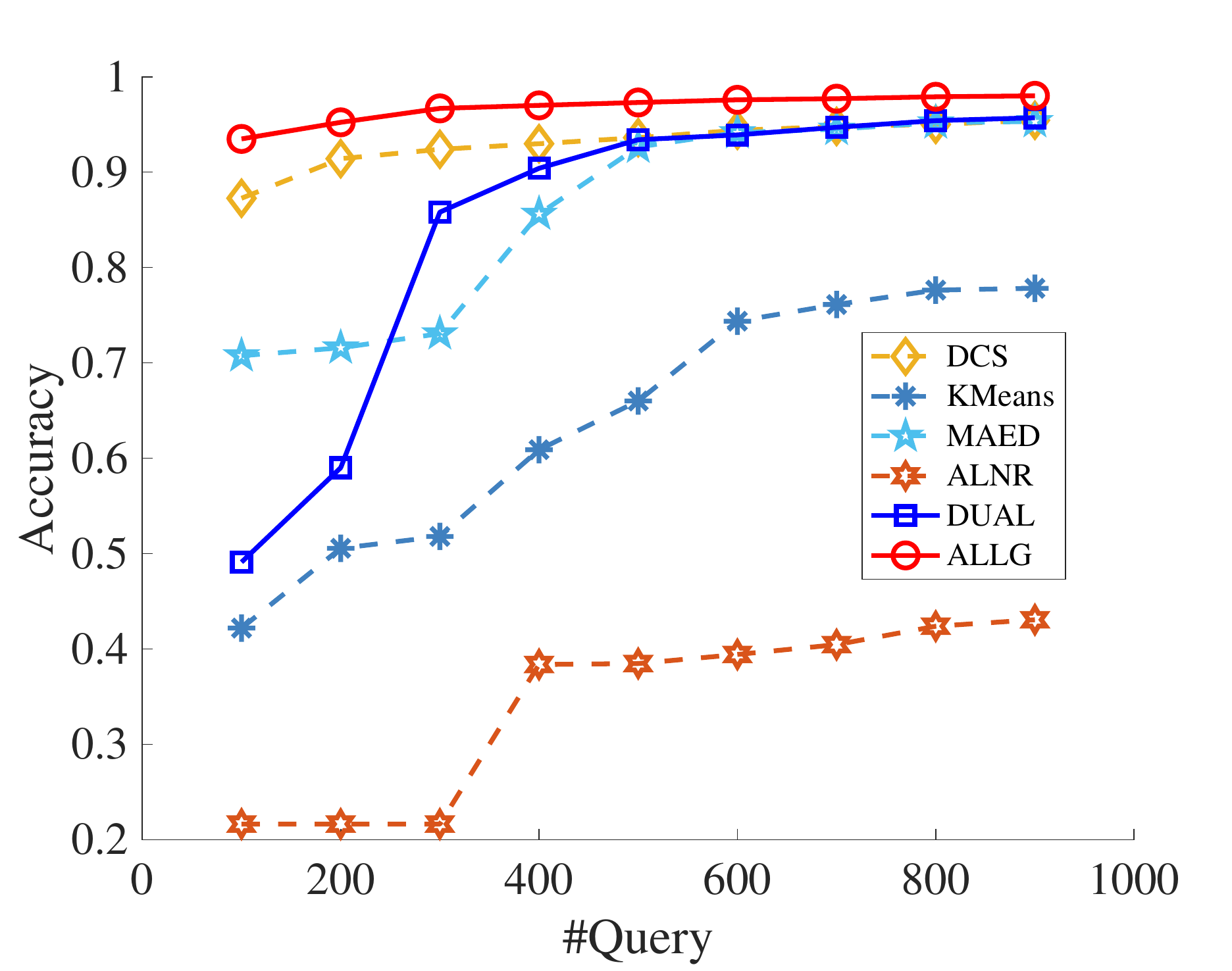}}
\subfigure[Letter Recognition]{\includegraphics[width=2.1in]{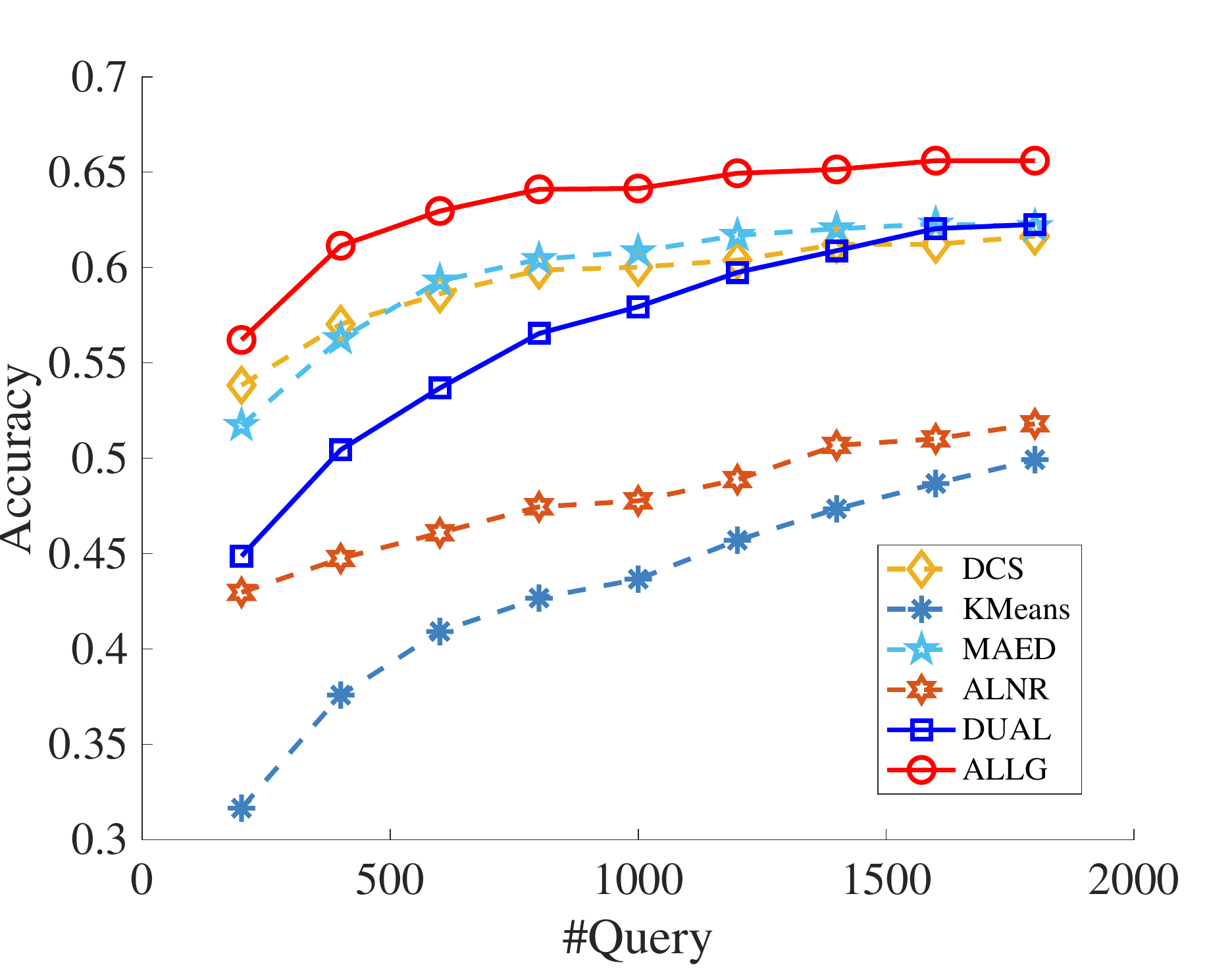}}
\caption{The performance comparisons of different active learning methods combined with SVM on six benchmark.}
\label{svm}
\end{figure*}

\begin{table}[t!]
\small
\centering
\begin{threeparttable}
\begin{tabular}{|c|c|c|c|}
\hline Datasets & Size & Dimension & Class \\
\hline \hline Splice-junction & 1000 & 60 & 2 \\
\hline Plant Species Leaves & 1600 & 64 & 100 \\
\hline Waveform & 5000 & 40 & 3 \\
\hline ESR & 11500 & 178 & 5 \\
\hline GSAD & 13910 & 128 & 6 \\
\hline Letter Recognition & 20000 & 16 & 10 \\
\hline
\end{tabular}
\end{threeparttable}
\caption{Details of experimental datasets.}
\label{dataset}
\end{table}

\begin{figure*}
\centering
%\subfigtopskip=0.5pt %设置子图与上面正文或别的内容的距离
%\subfigbottomskip=0.5pt %设置第二行子图与第一行子图的距离，即下面的头与上面的脚的距离
%\subfigcapskip=-3pt %设置子图与子标题之间的距离
\subfigure[Splice-junction]{\includegraphics[width=2.1in]{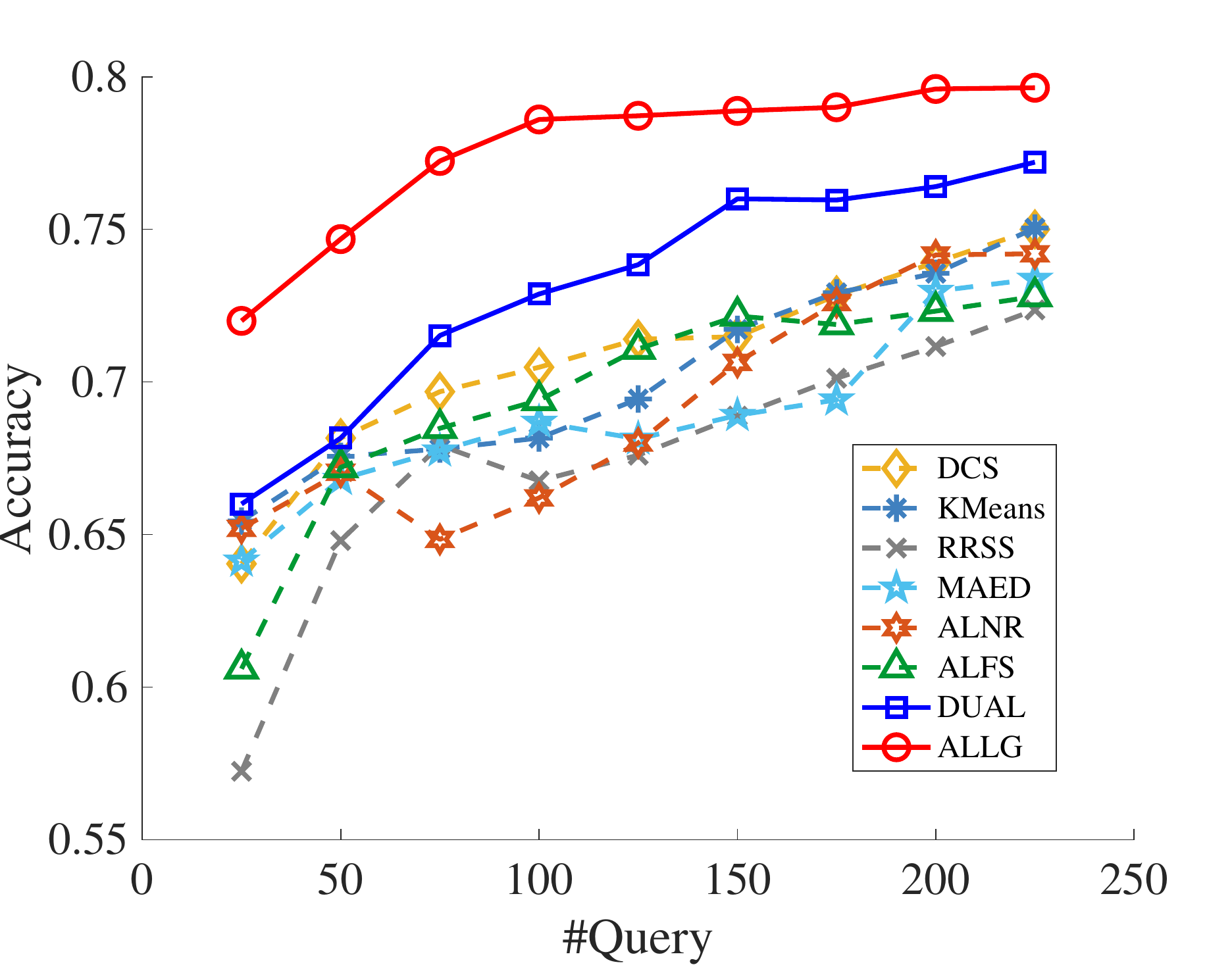}}
\subfigure[Plant Species Leaves]{\includegraphics[width=2.1in]{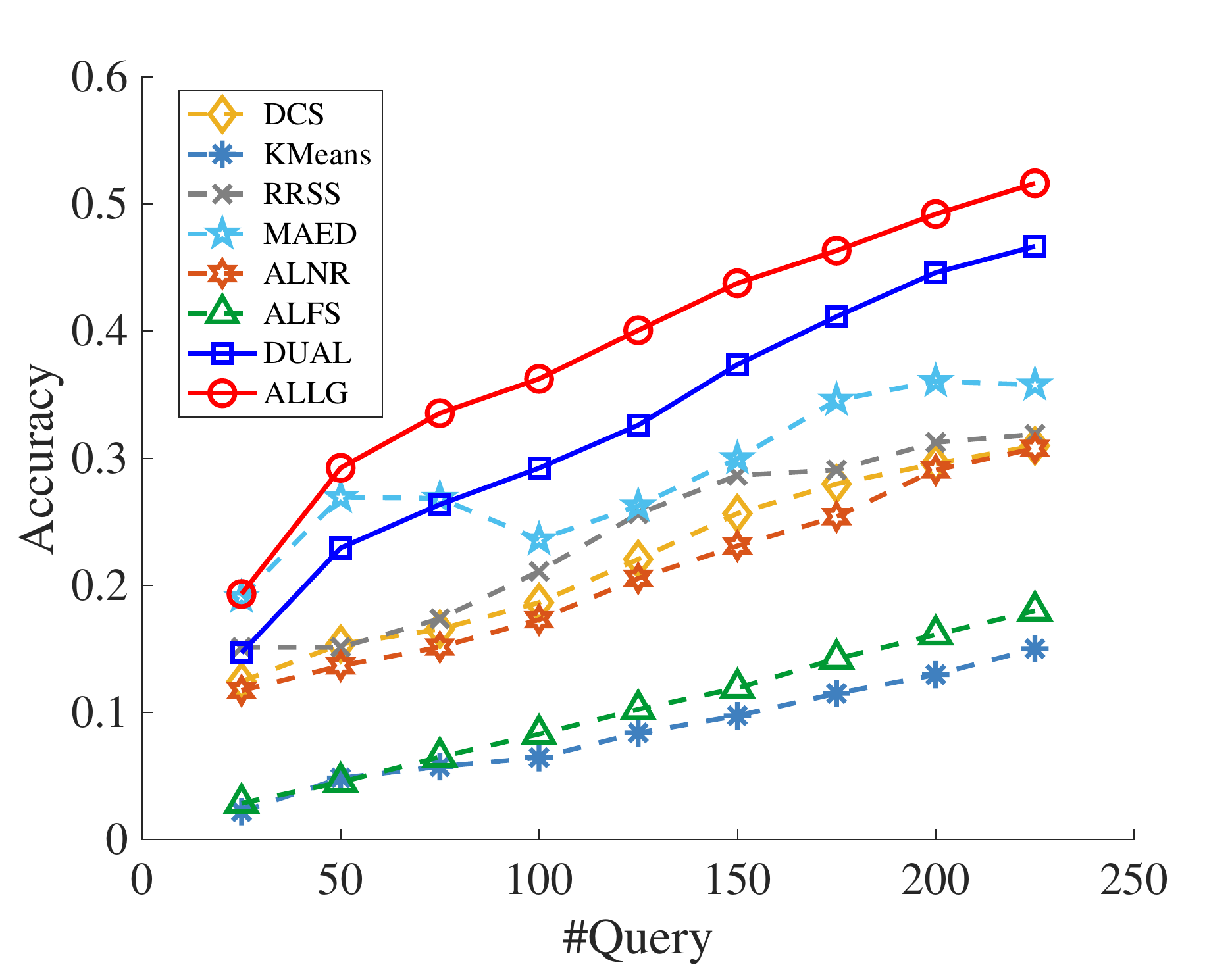}}
\subfigure[Waveform]{\includegraphics[width=2.1in]{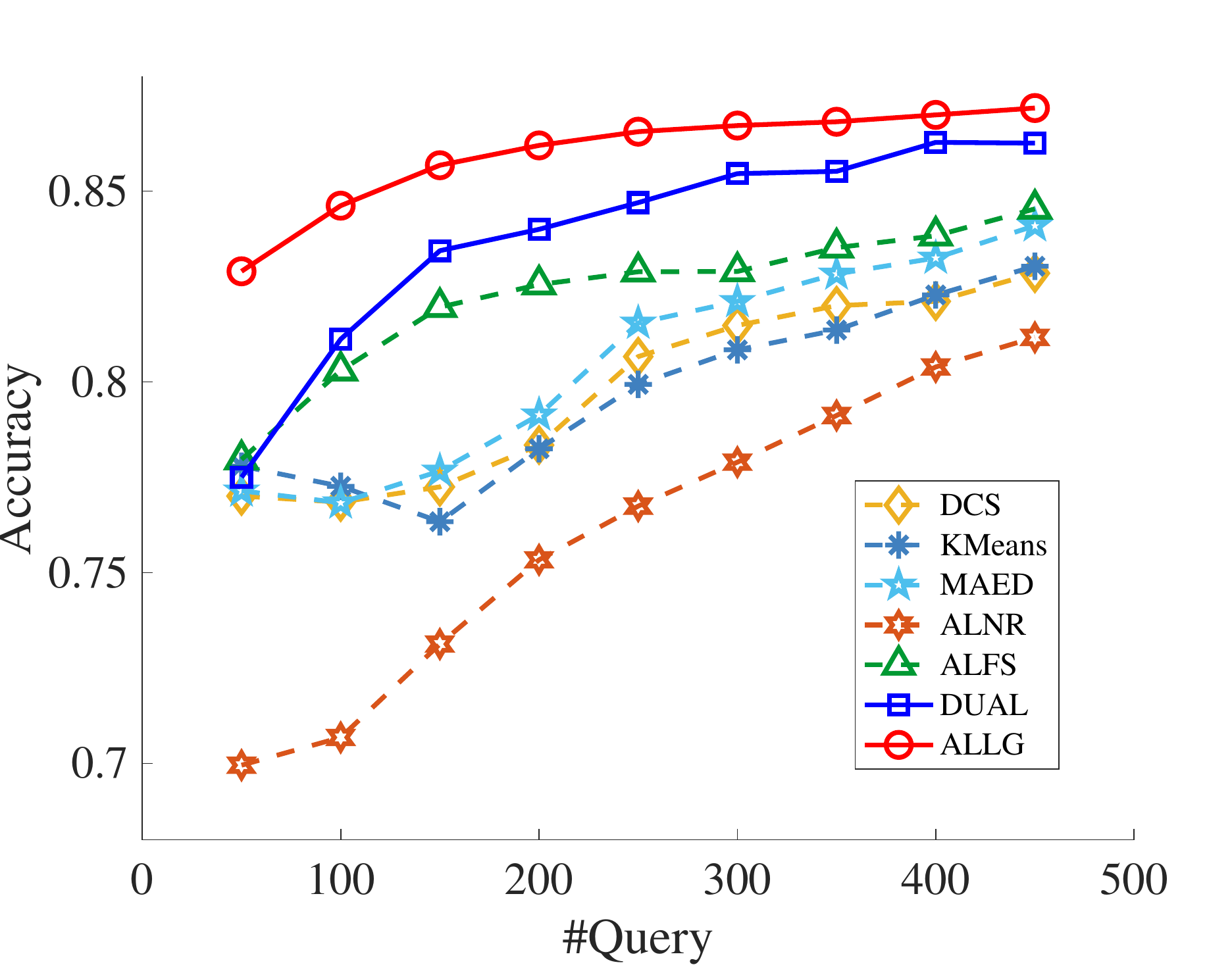}}
\subfigure[ESR]{\includegraphics[width=2.1in]{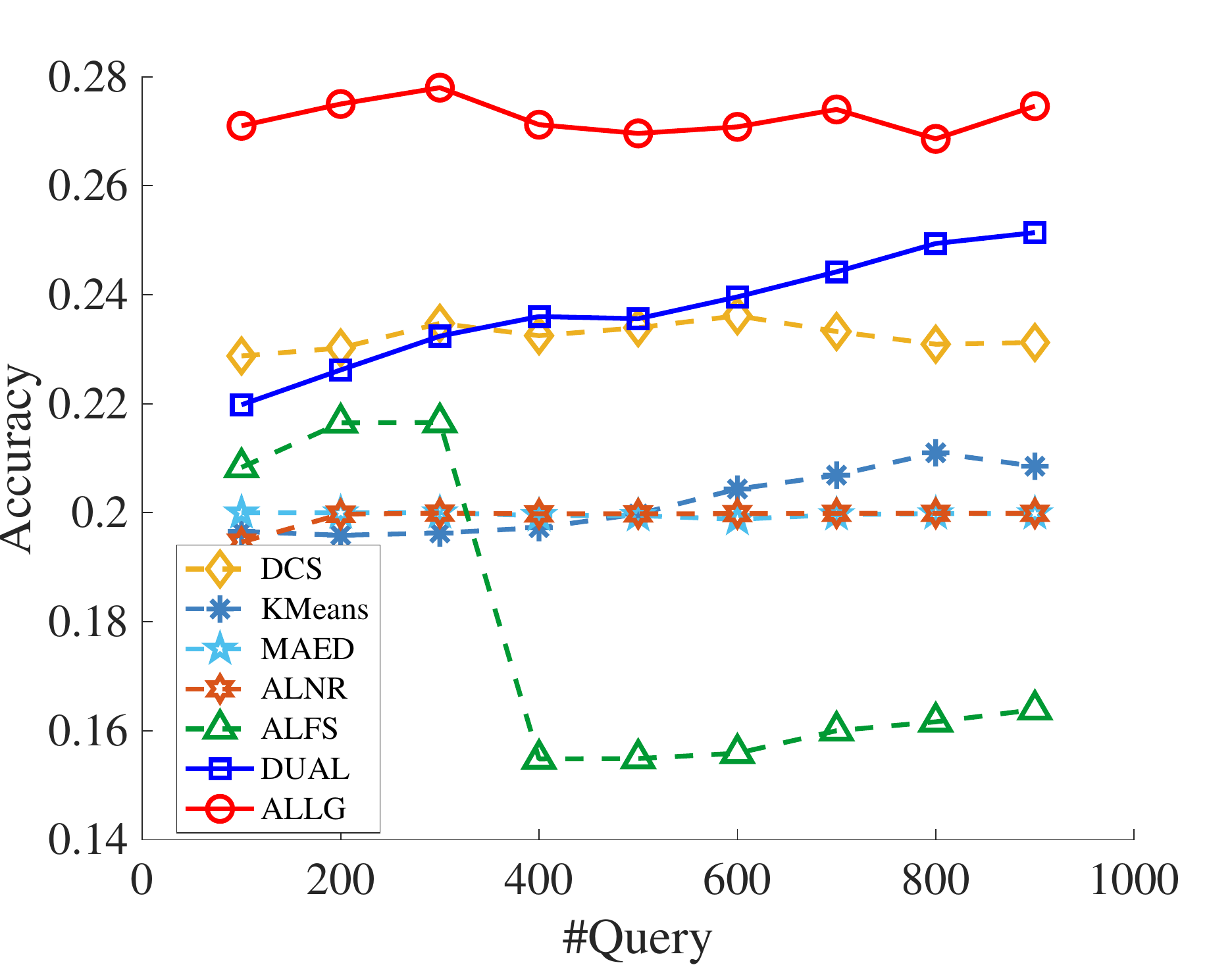}}
\subfigure[GSAD]{\includegraphics[width=2.1in]{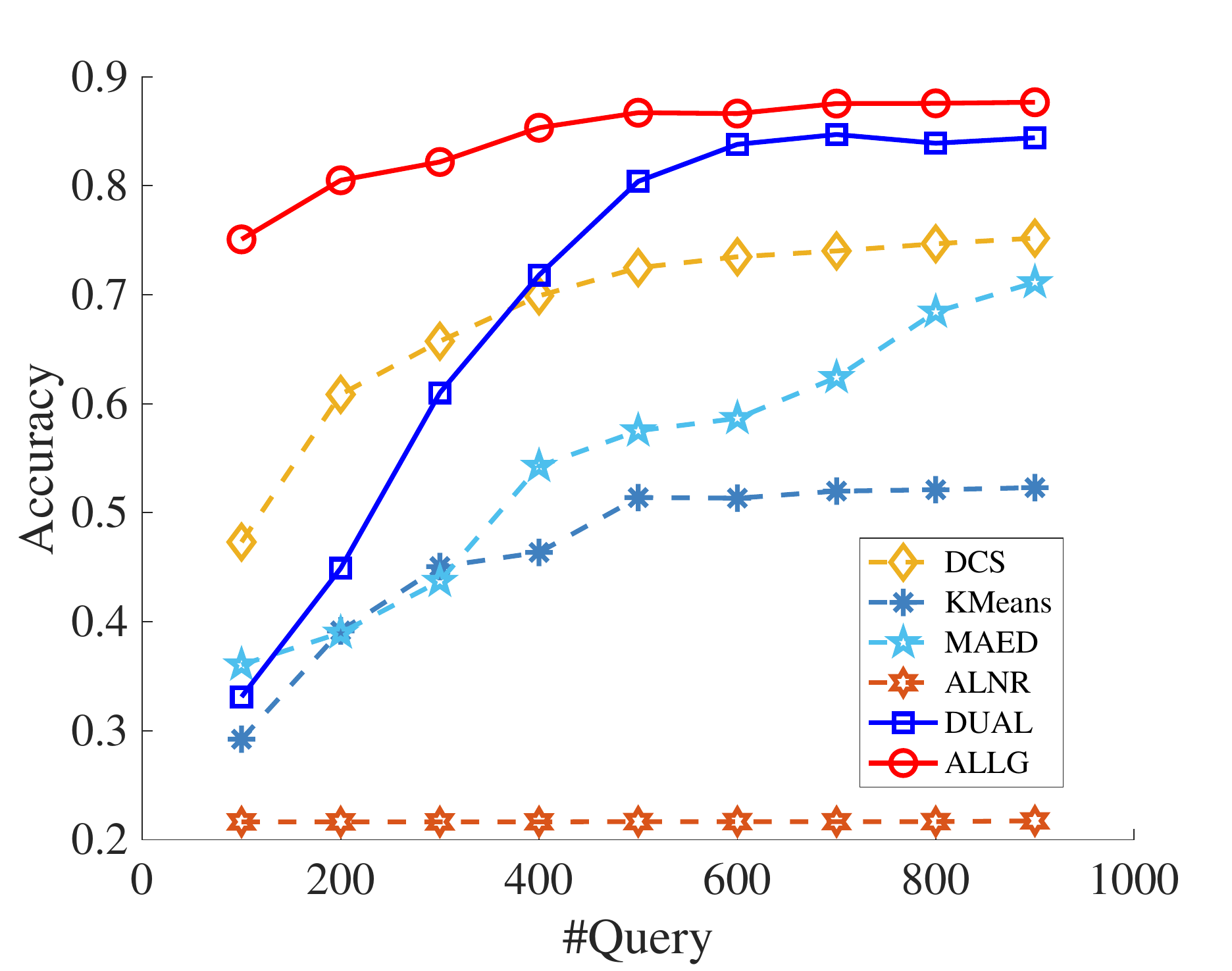}}
\subfigure[Letter Recognition]{\includegraphics[width=2.1in]{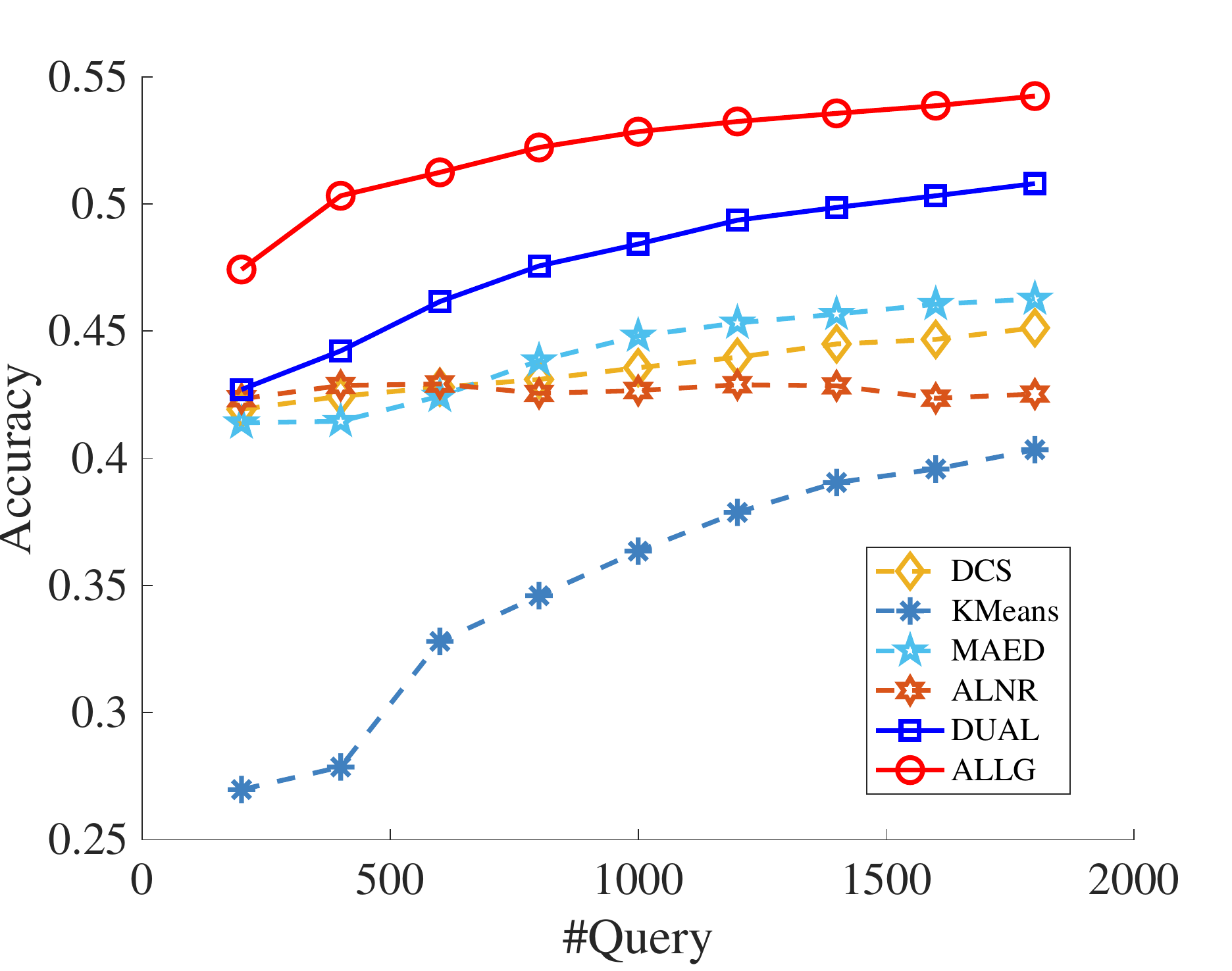}}
\caption{The performance comparisons of different active learning methods combined with Logistic Regression on six benchmark.}
\label{lr}
\end{figure*}

\begin{table*}[t!]
\small
\centering
\begin{threeparttable}
\begin{tabular}{c|ccccccccc|c}
\toprule
\# of Query & 25 & 50 & 75 & 100 & 125 & 150 & 175 & 200 & 225 & Average \\
\midrule
Kmeans&0.6616& 0.6856 &0.6692& 0.6688& 0.6820&  0.6884& 0.6960 & 0.6876&0.7080&0.6830\\
DCS&0.6360&  0.6720 & 0.6632& 0.6808 &0.6884 &0.6920  &0.6916& 0.7096& 0.7384&0.6857\\
RRSS&0.5844& 0.6516 &0.6712 &0.6564 &0.6680 & 0.6832& 0.6952 &0.7052 &0.7124&0.6697\\
MAED&0.6544 & 0.6720 &0.6709 &0.6725 &0.6641& 0.6632&  0.6689& 0.7116 & 0.7180&0.6772\\
ALNR&0.6472& 0.6552& 0.6396& 0.6368& 0.6744 &0.6824& 0.6992 &0.7272& 0.7336&0.6772\\
ALFS&0.6248 &0.6744 &0.6804& 0.6840 & 0.7072 &0.7104 &0.7136& 0.7284 &0.7416&0.6960\\
\hline
DUAL\_O&0.6552 &0.6740 & 0.6976& 0.7088& 0.7012 &0.7168& 0.7212& 0.7184& 0.7380&0.7034 \\
DUAL&0.6524 &0.6688 &0.7064 &0.7148& 0.7200 &  0.7456 &0.7504& 0.7600 &  0.7620&0.7200 \\
\hline
ALLG\_O&\textbf{0.7212} &\underline{0.7328}& \underline{0.7484} &\underline{0.7540} & \underline{0.7612}& \underline{0.7680}&\underline{ 0.7720} & \underline{0.7744} &\underline{0.7772}&\underline{0.7565}\\
ALLG &\underline{0.7132}& \textbf{0.7484} &\textbf{0.7640}  &\textbf{0.7660} & \textbf{0.7756}& \textbf{0.7848}& \textbf{0.7892}& \textbf{0.7948}& \textbf{0.7968} & \textbf{0.7703}\\
\bottomrule
\end{tabular}
\end{threeparttable}
\caption{Ablation study on sample representation. The bold text indicates the best results and the underlined text indicates the second best results.}
\label{ablation1}
\end{table*}

\subsection{Experimental Setting}
\noindent \textbf{Dataset:} 
To demonstrate that datasets from different domains can benefit from ALLG, we conduct experiments on six publicly available datasets from different domains, and the details of the datasets are summarized in Table \ref{dataset}\footnote{These datasets are all downloaded from the UCI Machine Learning Repository: https://archive.ics.uci.edu/ml/datasets.php.}.
In the datasets, "Splice-junction" and "ESR" are from biology, while "ESR" is present as time-series. 
"Plant Species Leaves" and "Letter Recognition" are generated from images. 
"Waveform" is a physical dataset, while "GSAD" is from sensors utilized in simulations.

\vspace{0.1in}
\noindent \textbf{Baseline\footnote{All source codes are obtained from the authors of the corresponding papers, except K-means and ALNR.}:}
We compare our method with several typical unsupervised active learning algorithms, including RRSS \cite{Nie2013Early}, ALNR \cite{hu2013active}, MAED \cite{cai2011manifold}, ALFS \cite{li2018joint}, DUAL \cite{liijcai20}. 
We also compare with a matrix column subset selection algorithm, deterministic column sampling (DCS) \cite{papailiopoulos2014provable}, which can be used for unsupervised active learning. 
In addition, we take $K$-Means as another baseline, in which we choose the samples which are closest to the cluster centers as the most representative samples, and we set $K=5$ in experiments.

\vspace{0.1in}
\noindent \textbf{Experimental protocol:} 
Following \cite{li2018joint} and \cite{liijcai20}, we randomly select 50\% of the samples as the candidate set and the rest is testing set.
Different active learning algorithms are performed on the same candidate set to query the most representative $m$ samples.
To verify the quality of samples selected by these methods, we train two classifiers by using these selected samples as the training data: a SVM classifier with a linear kernel and C = 100, as well as a Logistic Regression (LR) classifier. 
We search the trade-off parameters in our algorithm from $\{0.1, 1, 10\}$. For a fair comparison, the parameters of RRSS, ALNR, MAED, ALFS, DUAL are all searched from the same space. 
Each experiment is run five times, and the result is reported in terms of the average accuracy.

\begin{table*}[ht]
\small
\centering
\begin{tabular}{c|ccccccccc|c}
\toprule
\# of Query & 25 & 50 & 75 & 100 & 125 & 150 & 175 & 200 & 225 & Average\\
\midrule
w/o smoothing &0.6486 & 0.7126 & 0.7313 & 0.7320 & 0.7540 & 0.7540 & 0.7526 & 0.7606 & 0.7686 & 0.7349 \\
\hline
ALLG\_knn &0.6800 & 0.7160 & 0.7055 & 0.7225 & 0.7415 & 0.7535 & 0.7725 & 0.7695 & 0.7855 & 0.7384 \\
ALLG\_one &0.6900 & 0.7195 & 0.7385 & 0.7540 & 0.7730 &0.7805 & 0.7780 & 0.7820 & 0.7840 & 0.7555 \\
ALLG\_ts & 0.6820&  0.7430&  0.7380&  0.7430&  0.7540&  0.7820&  0.7890&  0.7840&  0.7850&0.7555\\
ALLG\_td &\textbf{0.7145} & 0.7430 & 0.7505 & 0.7630 & 0.7735 & 0.7715 & 0.7855 & \textbf{0.7970} & 0.7940 & 0.7658\\
\hline
ALLG & 0.7132 & \textbf{0.7484} & \textbf{0.7640} & \textbf{0.7660} & \textbf{0.7756} & \textbf{0.7848} & \textbf{0.7892} & 0.7948 & \textbf{0.7968} & \textbf{0.7703}\\
\bottomrule
\end{tabular}
\caption{Ablation study on graph structure learning. The bold text indicates the best results.}
\label{ablation2}
\end{table*}

\subsection{Experimental Result}
\noindent{\textbf{General Performance:}} Figure \ref{svm} and Figure \ref{lr} show the results of different methods combined with a SVM classifier and a LR classifier respectively.
We can observe that ALLG outperforms all other baselines in almost all queries. We use two different classifiers to illustrate that the quality of selected samples by ALLG is agnostic to classifiers. 
It is worth noting that ALLG achieves marked improvement compared to deep learning based DUAL, about 3\% average improvement on different numbers of query.
It's verified that learning graphs of data can really be positive to sample selection.
Note that we do not perform ALFS and RRSS on larger datasets, because of their unaffordable computational complexities.

\vspace{0.1in}
\noindent{\textbf{Ablation Study:}} We perform ablation study on the Splice-junction dataset to gain further understanding of the proposed method. The experimental setting is as:
\begin{itemize}
\item \textbf{Training Classifiers Using the Original Features}: For ALLG and DUAL, they embed samples to a latent space, and use the new representation to train classifiers. 
To eliminate the influence of new representations, we use original features to train classifiers after obtaining the selected samples by ALLG and DUAL. We denote them as ALLG\_O and DUAL\_O respectively. 
The results are shown in Table \ref{ablation1}. In general, ALLG\_O and DUAL\_O still achieve better performance than other methods. Meanwhile, ALLG\_O has a better result than DUAL\_O, which verifies the effectiveness of our method once again. 
In addition, ALLG is better than ALLG\_O, and DUAL achieves better performance than DUAL\_O. This illustrates that learning a nonlinear representation can be helpful for active learning.

\item \textbf{Adjacent Matrices Learning Module}: 
We also verify the effectiveness of the graph structure learning module, i.e., adjacent matrix learning.
We denote: 
1) w/o smoothing: ALLG trained without graph structure learning; 
2) ALLG\_knn: ALLG using $k$-nearest neighbor graph as the only adjacent matrix; 
3) ALLG\_one: ALLG trained with one learnt adjacent matrix; 
4) ALLG\_ts: ALLG trained with two learnt adjacent matrices but forcing them to be the same;
5) ALLG\_td: ALLG trained with two different learnt adjacent matrices (i.e., it will become ALLG when shortcut connections are added).
%ALLG trained without graph structure learning as w/o smoothing; The ALLG using $k$-nearest neighbor graph as the final adjacent matrix is denoted as ALLG\_{knn}; ALLG\_1 denotes the one learned with an optimal adjacent matrix; The ALLG trained by learning two same adjacent matrices is denoted as ALLG\_{same}; Finally, the ALLG  trained by learning two different adjacent matrices is denoted as ALLG\_2;
The results are reported in Table \ref{ablation2}. 
We find that ALLG\_knn achieves better results than w/o smoothing, illustrating that leveraging graph structure of data is good for learning a better sample representation. 
ALLG\_one is better than ALLG\_knn, demonstrating that learning a graph structure can achieve superior results, compared to a human estimated one. ALLG\_td outperforms ALLG\_one and ALLG\_ts. This illustrates that learning multiple different adjacent matrices can be beneficial to representation learning. Finally, ALLG beats ALLG\_td, showing that shortcut connection is effective for unsupervised active learning.
\end{itemize}

\begin{figure}[H]
\centering
%\subfigure{\includegraphics[width=0.35\linewidth]{./figures/lambda.pdf}}
%\subfigure{\includegraphics[width=0.35\linewidth]{./figures/alpha.pdf}}
%\subfigure{\includegraphics[width=0.35\linewidth]{./figures/beta.pdf}}
\includegraphics[width=1\linewidth]{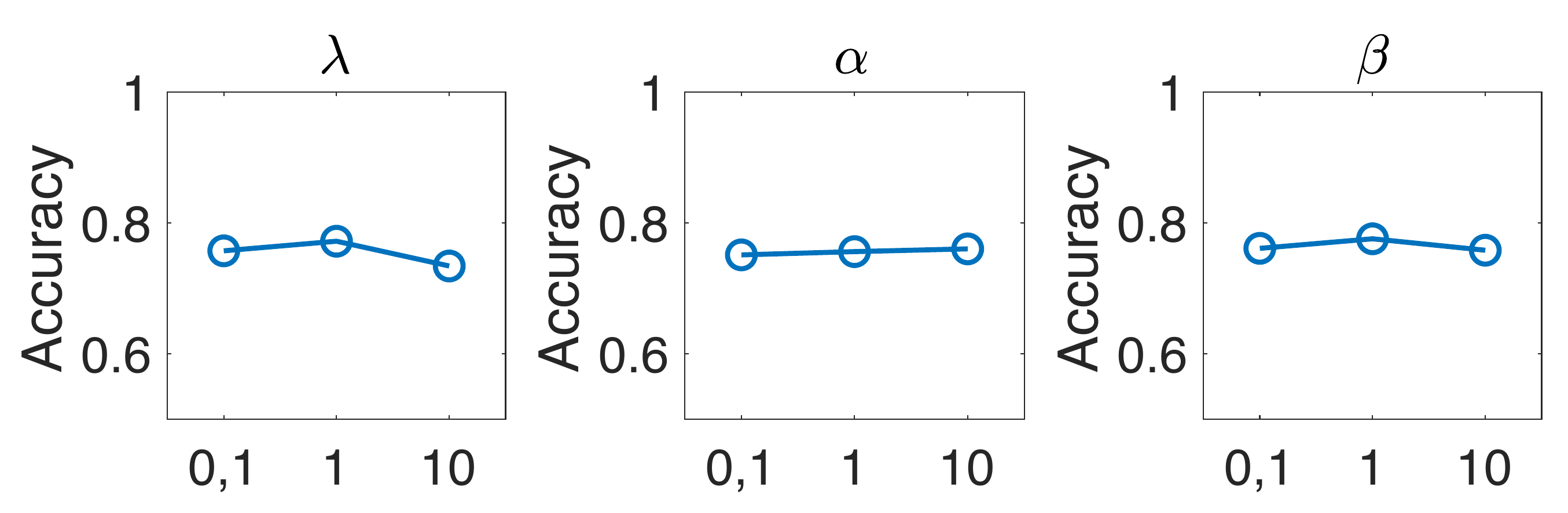}
\caption{Parameter study on Splice-junction dataset.}
\label{parameterstudy}
\end{figure}

\vspace{0.1in}
\noindent\textbf{Parameter Study:}
We study the sensitivity of our algorithm in terms of the trade-off parameters $\lambda, \alpha, \beta$ on the Splice-junction dataset. We fix the number of queries to 125. The results are shown in Figure \ref{parameterstudy}. Our method is insensitive to the parameters with a relatively wide range.

\begin{figure}[H]
\centering
\subfigure[Total loss]{\includegraphics[width=1.5in]{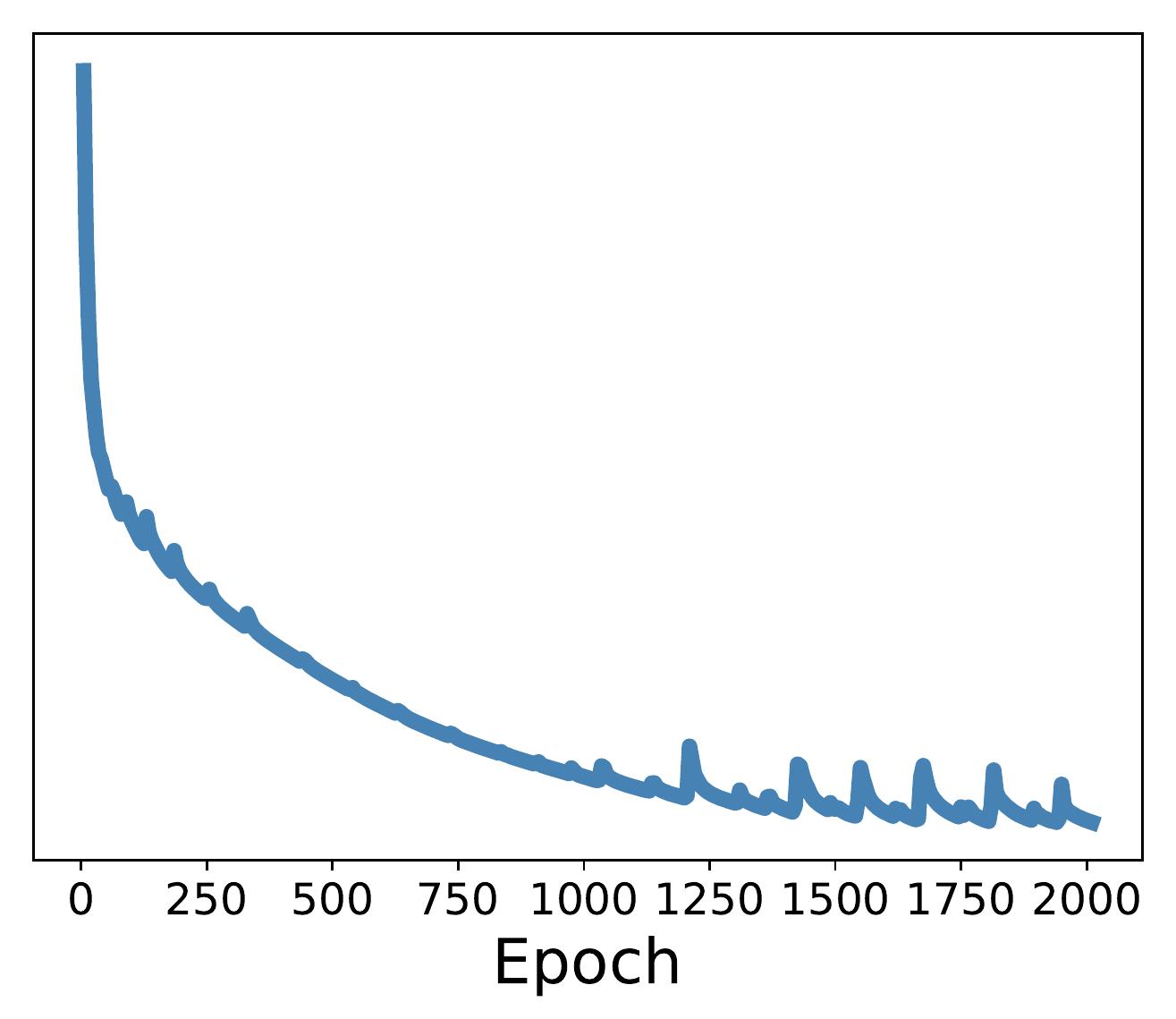}}
\subfigure[Supnorm loss of $\mathbf Q$]{\includegraphics[width=1.5in]{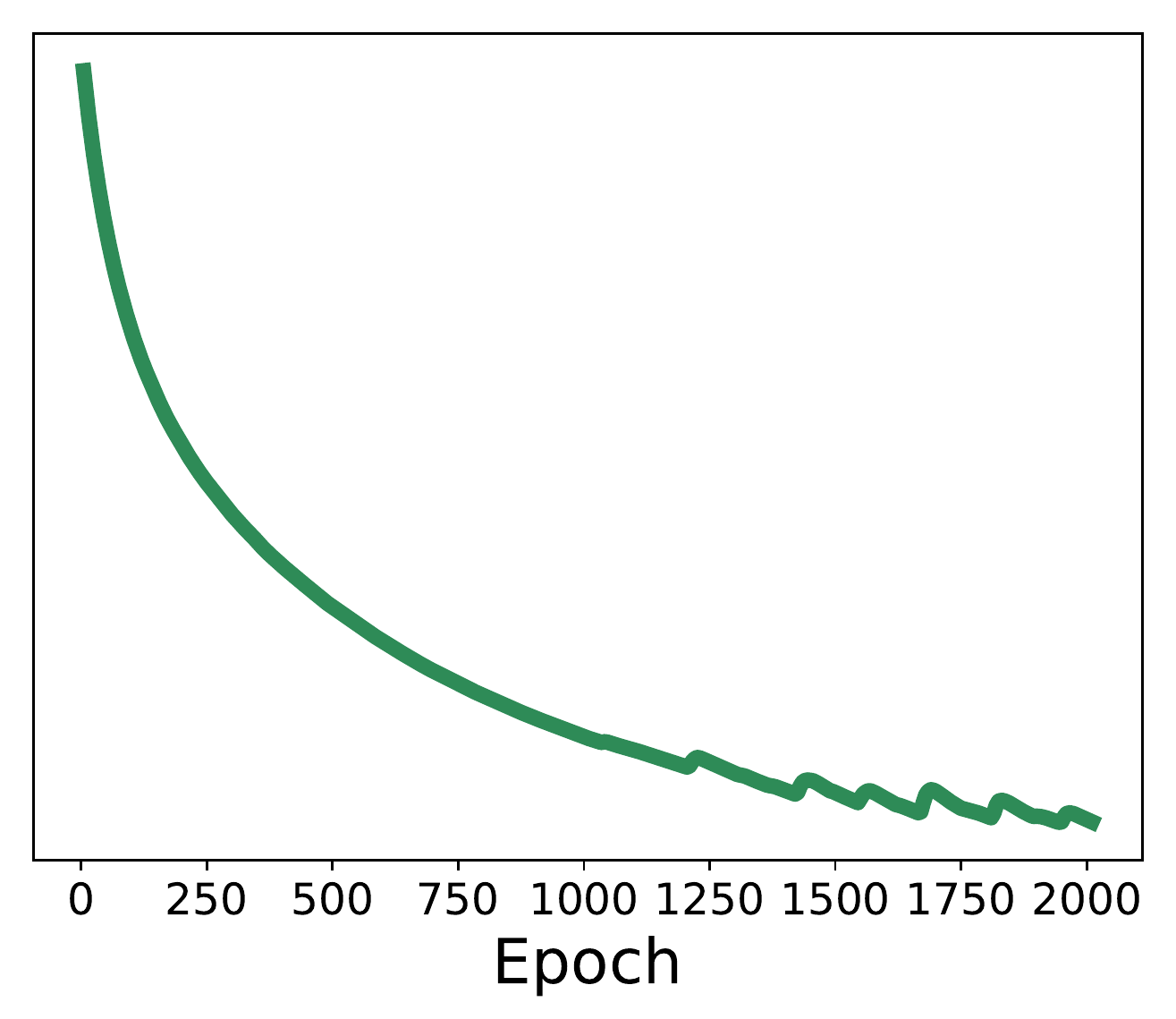}}
\caption{Convergence analysis of ALLG}
\label{loss}
\end{figure}

\vspace{0.1in}
\noindent \textbf{Convergence Analysis:}
%The convergence curves of ALLG are shown in Figure \ref{loss}. As the number of epoch increases, the total loss and the supnorm loss of $\mathbf Q$ are gradually decreased until convergent. On the Splice-junction dataset, the algorithm converges when the epoch reaches to about 2000.
We further show the convergence curves of ALLG on the Splice-junction dataset. The results are shown in Figure \ref{loss}. As the number of epoch increases, the total loss and the supnorm loss of $\mathbf Q$ are gradually decreased until convergent. From the figure, the algorithm converges when the epoch reaches to around 2000.

\begin{figure}[H]
\centering
\includegraphics[width=2.2in]{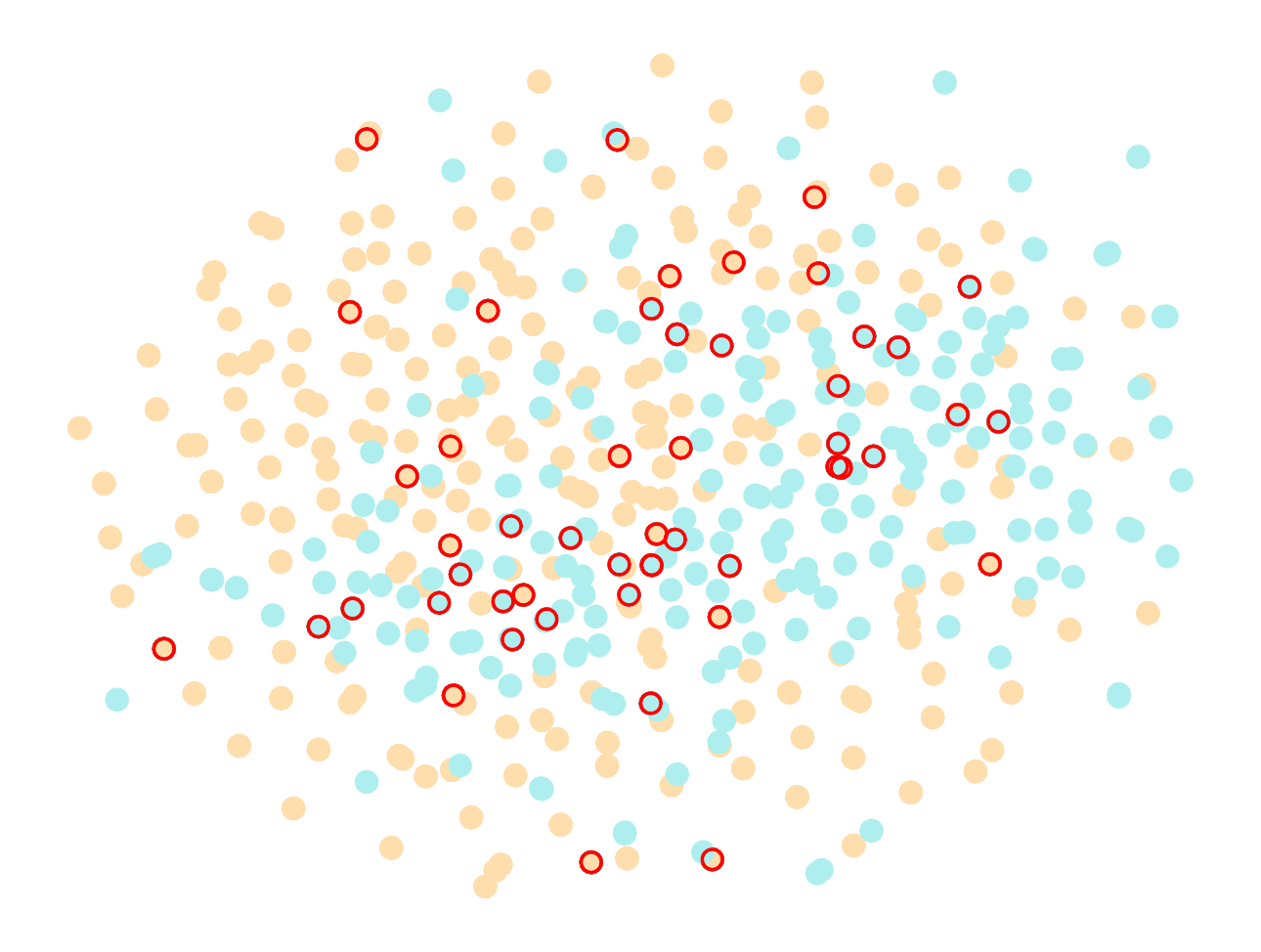}
\caption{Visualization by t-SNE on Splice-junction dataset. The red circles denote the selected samples, and other color solid circles denote different classes.}
\label{vis}
\end{figure}

\vspace{0.1in}
\noindent\textbf{Visualization:}
We use t-SNE \cite{maaten2008visualizing} to visualize  sample selection on the Splice-junction dataset shown in Figure\ref{vis}, the selected samples are distributed uniformly, and can better represent the whole dataset.

\section{Conclusion}
%In this paper, we proposed a novel unsupervised active learning model by learning optimal graph structures of data. 
%A novel adjacent matrices learning module was devised to learn a series of optimal graph structures of data through relation propagation regularization without pre-defining the graph structures.
%We took $k$-nearest neighbor graph prior to assist in learning. Besides, we utilized a shortcut connection to alleviate the over-smoothing problem. Experimental results demonstrated the effectiveness of the proposed method.

In this paper, we proposed a novel unsupervised active learning model by learning optimal graph structures of data. 
A novel adjacent matrices learning module was devised to learn a series of optimal graph structures of data through relation propagation regularization without pre-defining the graph structures. 
In the meantime, we took k-nearest neighbor graph prior to assist in graph structure learning. Finally, we utilized a shortcut connection to alleviate the over-smoothing problem. Experimental results demonstrated the effectiveness of our method.

% Use \bibliography{yourbibfile} instead or the References section will not appear in your paper

\newpage
\bibliographystyle{named}
\bibliography{aaai22}

\end{document}